\pdfoutput=1

\documentclass[11pt]{article}

\usepackage{acl}
\usepackage[ruled,vlined]{algorithm2e}
\usepackage{times}
\usepackage{latexsym}
\usepackage{graphicx}

\usepackage[T1]{fontenc}

\usepackage[utf8]{inputenc}

\usepackage{microtype}
\usepackage{booktabs}
\usepackage{amsmath}
\usepackage{soul}
\soulregister\cite7 
\soulregister\citep7 
\soulregister\citet7 
\soulregister\ref7 
\soulregister\pageref7 

\title{Multi-stage Distillation Framework for Cross-Lingual Semantic Similarity Matching}

\author{
Kunbo Ding\textsuperscript{\rm 1 \thanks{$^*$Contribution during internship at Tencent Inc.}}, 
Weijie Liu\textsuperscript{\rm 1,2}, 
Yuejian Fang\textsuperscript{\rm 1}, 
Zhe Zhao\textsuperscript{\rm 2}, 
Qi Ju\textsuperscript{\rm 2},
Xuefeng Yang\textsuperscript{\rm 2}\\
\bf{Rong Tian\textsuperscript{\rm 2},
Tao Zhu\textsuperscript{\rm 2},
Haoyan Liu\textsuperscript{\rm 2},
Han Guo\textsuperscript{\rm 2},
Xingyu Bai\textsuperscript{\rm 2},
Weiquan Mao\textsuperscript{\rm 2}}\\
\bf{Yudong Li\textsuperscript{\rm 2},
Weigang Guo\textsuperscript{\rm 2},
Taiqiang Wu\textsuperscript{\rm 2},
and Ningyuan Sun\textsuperscript{\rm 2}}\\ 
\textsuperscript{\rm 1}Peking University, Beijing, China  \textsuperscript{\rm 2}Tencent Research, Beijing, China\\
\footnotesize{kunbo\_ding@stu.pku.edu.cn, dataliu@pku.edu.cn, fangyj@ss.pku.edu.cn, \{nlpzhezhao, damonju, ryanxfyang\}@tencent.com}\\
\footnotesize{\{rometian, mardozhu, haoyanliu, serenguo, cynthyabai, weiquanmao, yudongli, jimwgguo, takiwu, waynenysun\}@tencent.com}\\
}

\begin{document}
\maketitle
\begin{abstract}

Previous studies have proved that cross-lingual knowledge distillation can significantly improve the performance of pre-trained models for cross-lingual similarity matching tasks. However, the student model needs to be large in this operation. Otherwise, its performance will drop sharply, thus making it impractical to be deployed to memory-limited devices. To address this issue, we delve into cross-lingual knowledge distillation and propose a multi-stage distillation framework for constructing a small-size but high-performance cross-lingual model. In our framework, contrastive learning, bottleneck, and parameter recurrent strategies are combined to prevent performance from being compromised during the compression process. The experimental results demonstrate that our method can compress the size of XLM-R and MiniLM by more than 50\%, while the performance is only reduced by about 1\%.

\end{abstract}

\section{Introduction}

On the internet, it is widespread to store texts in dozens of languages in one system. Cross-lingual similar text matching in multilingual systems is a great challenge for many scenarios, e.g., search engines, recommendation systems, question-answer robots, etc. \citep{ceretal2017semeval, hardalovetal2020exams, asaietal2021xor}. 

\begin{figure}[t]
	\centering
	\includegraphics[width=1.0\columnwidth]{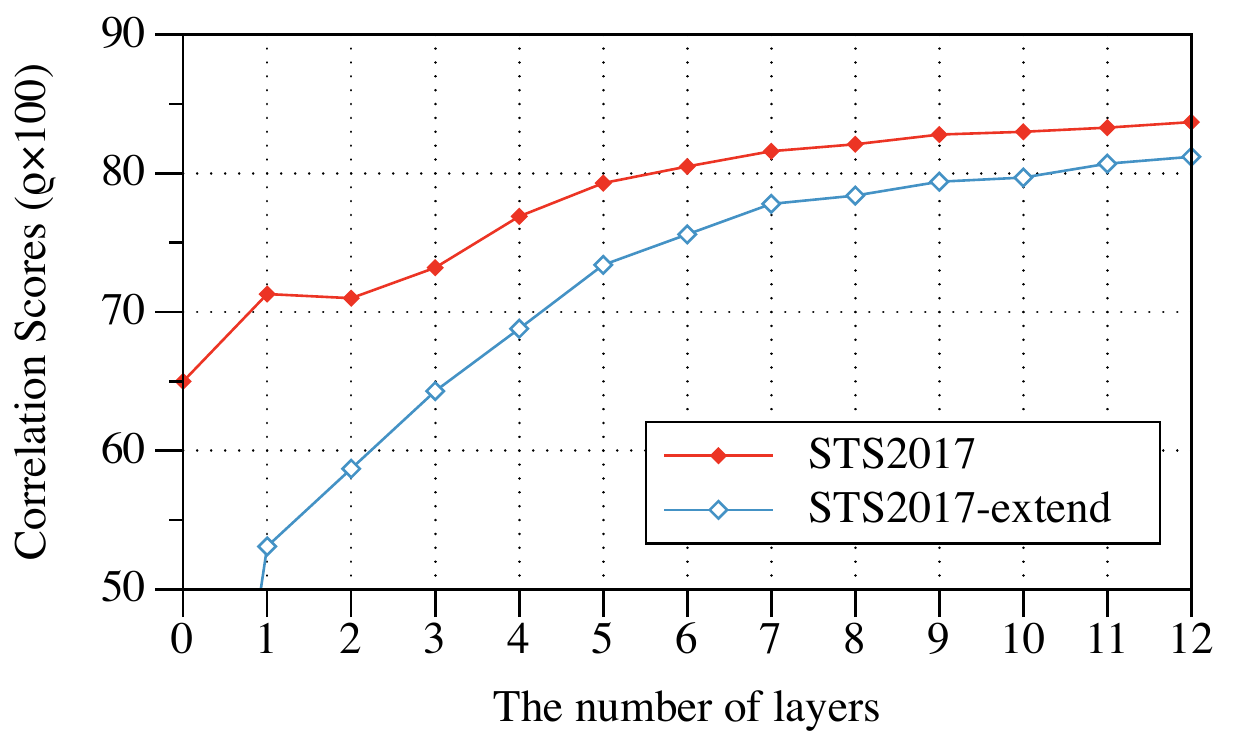}
	\caption{Evaluation results of XLM-R with different number of encoder layers on the STS2017 \textbf{monolingual} task and the STS2017-extend \textbf{cross-lingual} task, using SBERT-paraphrases for knowledge distillation.}

	\label{sts2017_intro}
\end{figure}

In the monolingual scenario, benefiting from the robust performance of the pre-trained language models (PLMs) (e.g., BERT \citep{devlin2019bert}, RoBERTa \citep{liu2019roberta}, T5 \citep{raffel2019exploring}, etc.), significant success has been achieved in text-similarity matching tasks. For example, \citet{reimersgurevych2019sentence} proposed the SBERT model trained with similar text pairs and achieved the state-of-the-art performance in the supervised similarity matching. In unsupervised scenarios, \citet{gaoetal2021simcse} proposed the SimCSE model, which was trained on Wiki corpus through contrastive learning task.

Drawing on the success in the monolingual scenario, researchers began to introduce pre-training technology into cross-lingual scenarios and proposed a series of multilingual pre-trained models, e.g., mBERT \citep{devlin2019bert}, XLM \citep{conneau2019cross}, XLM-R \citep{conneauetal2020unsupervised}, etc. Due to the vector collapse issue \citep{lietal2020sentence}, the performances of these cross-lingual models on similarity matching tasks are still not satisfactory. \citet{reimersgurevych2020making} injected the similarity matching ability of SBERT into the cross-lingual model through knowledge distillation, which alleviated the collapse issue and improved the performance of cross-lingual matching tasks.

Although the cross-lingual matching tasks have achieved positive results, the existing cross-lingual models are huge and challenging to be deployed in devices with limited memory. We try to distill the SBERT model into an XLM-R with fewer layers following \citet{reimersgurevych2020making}. However, as shown in Figure \ref{sts2017_intro}, the performance will be significantly reduced as the number of layers decreases. This phenomenon indicates that cross-lingual capabilities are highly dependent on the model size, and simply compressing the number of layers will bring a serious performance loss.

In this work, we propose a multi-stage distillation compression framework to build a small-size but high-performance model for cross-lingual similarity matching tasks. In this framework, we design three strategies to avoid semantic loss during compression, i.e., multilingual contrastive learning, parameter recurrent, and embedding bottleneck. We further investigate the effectiveness of the three strategies through ablation studies. Besides, we respectively explore the performance impact of reducing the embedding size and encoder size. Experimental results demonstrate that our method effectively reduces the size of the multilingual model with minimal semantic loss. Finally, our code
is publicly available\footnote{https://github.com/KB-Ding/Multi-stage-Distillaton-Framework}.

The main contributions of this paper can be summarized as follows:
\begin{itemize}
\item We validate that cross-lingual capability requires a larger model size and explore the semantic performance impact of shrinking the embedding or encoder size.
\item A multi-stage distillation framework is proposed to compress the size of cross-lingual models, where three strategies are combined to reduce semantic loss.
\item Extensive experiments examine the effectiveness of these three strategies and multi-stages used in our framework.

\end{itemize}

\section{Related work}

\begin{figure*}[t]
	\centering
	\includegraphics[width=1.0\linewidth]{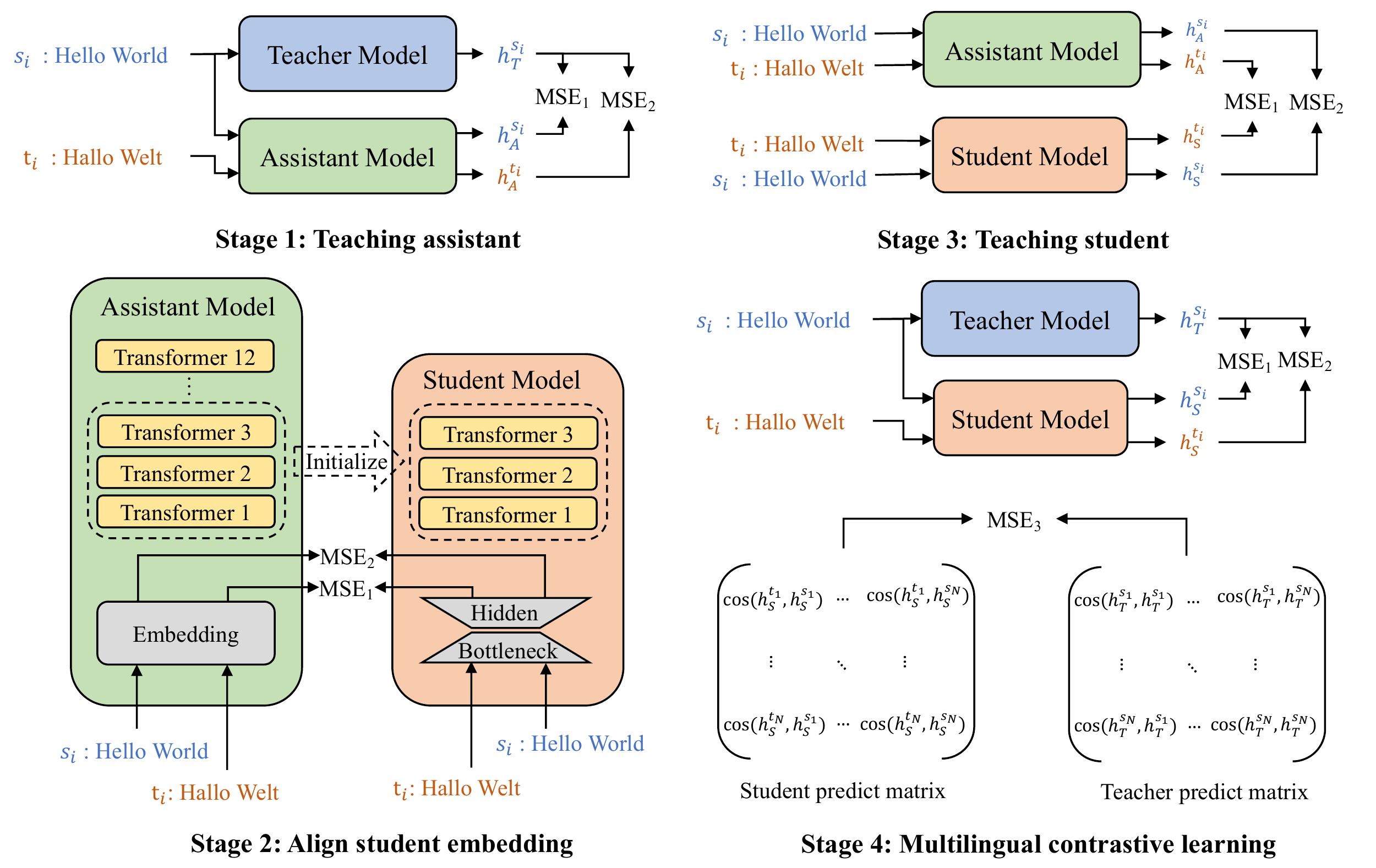}
	\caption{The overview of the model architecture and the multi-stage distillation. It consists of four stages and aims to obtain a small multilingual student model. For convenience, we take the English SBERT as the teacher model, XLM-R as the assistant model. $<s_{i},t_{i}>$ is a pair of parallel sentences in two different language. N is the batch size. MSE is the mean squared error loss function. }
	\label{architecture}
\end{figure*}

\subsection{Multilingual models}

Existing multilingual models can be divided into two categories, namely Multilingual general model and Cross-lingual representation model.

In the first category, transformer-based pre-trained models have been massively adopted in multilingual NLP tasks \citep{huangetal2019unicoder, chietal2021infoxlm, luoetal2021veco, ouyangetal2021ernie}. mBERT \citep{devlin2019bert} was pre-trained on Wikipedia corpus in 104 languages, achieved significant performance in the downstream task. XLM \citep{conneau2019cross} presented the translation language modeling (TLM) objective to improve the cross-lingual transferability by leveraging parallel data. XLM-R \citep{conneauetal2020unsupervised} was built on RoBERTa \citep{liu2019roberta} using CommonCrawl Corpus.

In the second category, LASER \citep{artetxeschwenk2019massively} used an encoder-decoder architecture based on a Bi-LSTM network and was trained on the parallel corpus obtained by neural machine translation. Multilingual Universal Sentence Encoder (mUSE) \citep{chidambarametal2019learning, yangetal2020multilingual} adopted a bi-encoder architecture and was trained with an additional translation ranking task. LaBSE \citep{feng2020language} turned the pre-trained BERT into a bi-encoder mode and was optimized with the objectives of mask language model (MLM) and TLM. Recently, \citet{maoetal2021lightweight} presented a lightweight bilingual sentence representation method based on the dual-transformer architecture.

\subsection{Knowledge distillation}

However, Multilingual models do not necessarily have cross-lingual capabilities, especially in the first category, in which vector spaces of different languages are not aligned. Knowledge distillation \citep{hinton2015distilling} used knowledge from a teacher model to guide the training of a student model, which can be used to compress the model and align its vector space at the same time.

For model compression, knowledge distillation aimed to transfer knowledge from a large model to a small model. BERT-PKD \citep{sunetal2019patient} extracted knowledge from both last layer and intermediate layers at fine-tuning stage. DistilBERT \citep{sanh2019distilbert} performed distillation at pre-training stage to halve the depth of BERT. TinyBERT \citep{jiaoetal2020tinybert} distilled knowledge from BERT at both pre-training and fine-tuning stages. MobileBERT \citep{sunetal2020mobilebert} distilled bert into a model with smaller dimensions at each layer. MiniLM \citep{wangetal2021minilmv2} conducted deep self-attention distillation. 

Unlike previous works presenting general distillation frameworks, we focus on compressing multilingual pre-trained models while aligning their cross-lingual vector spaces. In addition, we take inspiration from \citet{reimersgurevych2020making}, which successfully aligned the vector space of the multilingual model through cross-lingual knowledge distillation (X-KD). Our framework combines the advantages of X-KD for aligning vectors and introduces three strategies and an assistant model to prevent performance from being compromised during compression.

\section{Method}

In this section, we will introduce our method in detail. First, we exhibit the model architecture, and then introduce the multi-stage distillation strategy for the model training. An overview of our approach is shown in Figure \ref{architecture}.

\subsection{Model architecture}

Given a large-size monolingual model as teacher $T$ and a small-size multilingual model as student $S$, our goal is to transfer semantic similarity knowledge from $T$ to $S$ and simultaneously compress the size of $S$ with $m$ parallel sentences $P = \{<s_{1},t_{1}>, <s_{2},t_{2}>,\cdots <s_{m},t_{m}>\}$.
\subsubsection{Teacher model}
In this work, we use SBERT \citep{reimersgurevych2019sentence} as the teacher model, which has been proven to perform well on monolingual semantic similarity tasks. SBERT adopts a siamese network structure to fine-tune a BERT \citep{devlin2019bert} encoder, and applies a mean pooling operation to its output to derive sentence embedding.

\subsubsection{Assistant model} 
\citet{mirzadeh2020improved} proved that when the gap between the student and teacher is large, the performance of the student model will decrease. We hope to get a small student model with cross-lingual capabilities, while the teacher is a large monolingual model. To alleviate the gaps, we introduce an assistant model $A$ \citep{mirzadeh2020improved}, which is a large multilingual model with cross-lingual ability. 

\subsubsection{Student model} Inspired by ALBERT \citep{lan2019albert}, we design the student model with Parameter Recurrent and Embedding Bottleneck strategy. Since there is no available multilingual ALBERT, we need to design from scratch.

\noindent\textbf{Parameter recurrent.} We choose the first $M$ layers of the assistant model as a recurring unit (RU). The role of RU is to initialize the student model with layers from the assistant model. Concretely, the RU is defined as,
\begin{equation}
\label{eq:RU}
RU = \left \{ L_i| i\in \left [ 1,M \right ] \right \},
\end{equation}
where $L_i$ is the $i^{th}$ transformer layer.

\noindent\textbf{Embedding bottleneck.} Multilingual pre-trained models usually require a large vocabulary $V$ to support more languages, which leads to large embedding layer parameters. We add a bottleneck layer \citep{he2016deep, lan2019albert,sunetal2020mobilebert} of size $B$ between embedding layer and hidden layer $H$. In this way, the embedding layer is reduced from $O(V\times H)$ to $O(V\times B + B\times H)$.

\subsection{Multi-stage distillation}

Multi-stage Distillation is the key for enabling the small-size student model with cross-lingual matching ability.

\subsubsection*{Stage 1. Teaching assistant}

As the \textbf{Stage 1} in Figure \ref{architecture}, we use the teacher model and parallel corpus to align vector space between different languages through the loss function in (\ref{eq:TA}), enabling its cross-lingual ability \citep{reimersgurevych2020making}.
\begin{equation}
\label{eq:TA}
\ell _{stage1} = \frac{1}{|N|} \sum_{i}^{N} \left[ (h_{T}^{si} - h_{A}^{si})^2 + (h_{T}^{si} - h_{A}^{ti})^2 \right],
\end{equation}
where $N$ is the batch size, and $s_i$ and $t_i$ denotes the parallel sentences in a mini batch.

\subsubsection*{Stage 2. Align student embedding}

As the \textbf{Stage 2} in Figure \ref{architecture}, we align the embedding bottleneck layer with the assistant embedding space through the loss function in (\ref{eq:TA_Embedding}),
\begin{equation}
\label{eq:TA_Embedding}
\ell _{stage2} = \frac{1}{|N|} \sum_{i}^{N} \left[ (h_{Ae}^{si} - h_{Be}^{si})^2 + (h_{Be}^{ti} - h_{Ae}^{ti})^2 \right],
\end{equation}
where $h_{Ae}^{si}, h_{Ae}^{ti}$ denotes the output of assistant embedding layer, $h_{Be}^{si}, h_{Be}^{ti}$ denotes the output of embedding bottleneck layer. 

\subsubsection*{Stage 3. Teaching student}

In the \textbf{Stage 3}, the student model is trained to imitate the output of the assistant model with loss function in (\ref{eq:TA_Student}),
\begin{equation}
\label{eq:TA_Student}
\ell _{stage3} = \frac{1}{|N|} \sum_{i}^{N} \left[ (h_{A}^{si} - h_{S}^{si})^2 + (h_{S}^{ti} - h_{A}^{ti})^2 \right],
\end{equation}
where $h_{A}^{si}, h_{A}^{ti}$ denotes the output of assistant model, $h_{S}^{si}, h_{S}^{ti}$ denotes the output of student model.

\begin{table*}[t]
	\centering 
	\scriptsize
	\setlength{\tabcolsep}{4mm}{
		\begin{tabular}{lcccccc}
        \toprule
		\textbf{Model} & \textbf{AR-AR} & \textbf{ES-ES} & \textbf{EN-EN}  & \textbf{Avg.} & \textbf{Embedding size} & \textbf{Encoder size} \\ 
		\midrule
		\multicolumn{6}{l}{\emph{Pre-trained Model}}\\
		\verb|mBERT(mean)|  & 50.9 & 56.7 & 54.4 & 54.0 & 92.20M & 85.05M \\ 
		\verb|XLM-R(mean)|  & 25.7 & 51.8 & 50.7 & 42.7 & 192.40M & 85.05M  \\ 
		\verb|mBERT-nli-stsb| & 65.3  & 83.9 & 80.2 & 76.5 & 92.20M & 85.05M  \\ 
		\verb|XLM-R-nli-stsb| & 64.4 & 83.1 & 78.2 & 75.3 & 192.40M & 85.05M \\
		\verb|LASER| & 68.9 & 79.7 & 77.6 & 75.4 & 23.56M & 17.06M \\
		\verb|LaBSE| & 69.1 & 80.8 & 79.4 & 76.4 & 385.28M & 85.05M \\
		\midrule
		\multicolumn{6}{l}{\emph{Knowledge Distillation}} \\ 
		\verb|mBERT|$\leftarrow$ \verb|SBERT-nli-stsb|  & 78.8 & 83.0 & 82.5 & 81.4 & 92.20M & 85.05M \\ 

		\verb|XLM-R|$\leftarrow$ \verb|SBERT-nli-stsb|  & 79.9 & 83.5 & 82.5 & 82.0 & 192.40M & 85.05M\\ 
		
		\verb|mBERT|$\leftarrow$ \verb|SBERT-paraphrases|  & 79.1  & 86.5 & 88.2 & 84.6 & 92.20M & 85.05M\\
		\verb|DistilmBERT|$\leftarrow$ \verb|SBERT-paraphrases|  & 77.7  & 85.8 & 88.5 & 84.0 & 92.20M & 46.10M\\
		
		\verb|XLM-R|$\leftarrow$ \verb|SBERT-paraphrases|  & 79.6 & 86.3 & 88.8 & \textbf{84.6} & 192.40M & 85.05M\\
		\verb|MiniLM|$\leftarrow$ \verb|SBERT-paraphrases|  & 80.3  & 84.9 & 85.4 & \textbf{83.5} & 96.21M & 21.29M\\ 
		\midrule
		\multicolumn{6}{l}{\emph{Ours(Teacher model=SBERT-paraphrases)}} \\ 
		\verb|XLM-R|($b=True$, $bs=128$, $\left | RU \right | = 3$)  & 76.7  & 84.5 & 86.6  & 82.6 & 32.49M & 21.26M\\ 
		\verb|XLM-R|($b=True$, $bs=128$, $\left | RU \right | = 12$)  & 79.0  & 85.5 & 88.4 & \textbf{84.3} & \textbf{32.49M} & 85.05M\\ 
		\verb|XLM-R|($b=False$, $\left | RU \right | = 3$)  & 79.9  & 86.8 & 88.4 & \textbf{85.0} & 192.40M & \textbf{21.26M}\\ 

        \midrule
        \multicolumn{6}{l}{\emph{Ours(Teacher model=SBERT-paraphrases)}} \\ 
		\verb|MiniLM|($b=True$, $bs=128$, $\left | RU \right | = 3$)  & 72.8  & 79.3 & 84.4 & 78.8 & 32.05M & 5.32M\\ 
		\verb|MiniLM|($b=True$, $bs=128$,$\left | RU \right | = 12$)  & 79.0  & 84.4 & 85.2 & \textbf{82.9} & \textbf{32.05M} & 21.29M\\ 
		\verb|MiniLM|($b=False$, $\left | RU \right | = 3$)  & 79.9  & 85.3 & 85.6 & \textbf{83.6} & 96.21M & \textbf{5.32M}\\ 

        \bottomrule

	\end{tabular}
	
	}
	\caption{Spearman rank correlation ($\rho \times 100$) between the cosine similarity of sentence representations and the gold labels for STS 2017 \textbf{monolingual} dataset. $b$ indicates whether to use the Embedding Bottleneck strategy, $bs$ indicates the hidden size of Bottleneck layer. $|RU|$ indicates the first $|RU|$ layers from the basic model are taken as Recurrent Unit, the recurrent times $=$ basic model layers$/|RU|$.}
	\label{table_sts2017_mono}
\end{table*}

\subsubsection*{Stage 4. Multilingual contrastive learning}
After the above three stages, we can get a small multilingual sentence embedding model. However, as shown in Figure \ref{sts2017_intro}, when the model size decrease, its cross-lingual performance decreases sharply. Therefore, in this stage, we propose multilingual contrastive learning (MCL) task further to improve the performance of the small student model. 

Assuming the batch size is $N$, for a specific translation sentence pair $(s_i, t_i)$ in one batch, the mean-pooled sentence embedding of the student model is $(h_{S}^{si}, h_{S}^{ti})$. The MCL task takes parallel sentence pair $(h_{S}^{si}, h_{S}^{ti})$ as positive one, and other sentences in the same batch $\left \{ (h_{S}^{si}, h_{S}^{tj})|j\in \left [ 1,N \right ], j\ne i \right \} $as negative samples. Considering that the MCL task needs to be combined with knowledge distillation. Unlike the previous work \citep{yang2019improving, feng2020language, maoetal2021lightweight}, the MCL task does not directly apply the temperature-scaled cross-entropy loss function. 

Here, we introduce the implementation of the MCL task. For each pair of negative examples $(s_i, t_j)$ in the parallel corpus, the MCL task first unifies $(s_i, t_j)$ into the source language $(s_i, s_j)$, then uses the fine-grained distance between $h_{T}^{si}$ and $h_{T}^{sj}$ in the teacher model to push away the semantic different pair $(h_{S}^{si}, h_{S}^{tj})$ in the student model. For positive examples, the MCL task pull semantically similar pair $(h_{S}^{si}, h_{S}^{ti})$ together. The MCL task loss is (\ref{eq:MCL}),

\begin{equation}
\label{eq:MCL}
\ell _{1}  = \frac{1}{N^{2} } \sum_{i}^{N}\sum_{j}^{N}  \left (\phi (h_{T}^{si},h_{T}^{sj}) -\phi (h_{S}^{si},h_{S}^{tj})  \right ) ^{2},
\end{equation}

where $\phi$ is the distance function. Following prior work \citep{yang2019improving, feng2020language}, we set $\phi(x,y)=cosine(x,y)$. we also add the knowledge distillation task for multilingual sentence representation learning. The knowledge distillation loss is defined as,

\begin{equation}
\label{eq:Stu_mkd}
\ell _{2} = \frac{1}{|N|} \sum_{i}^{N} \left[ (h_{T}^{si} - h_{S}^{si})^2 + (h_{T}^{si} - h_{S}^{ti})^2 \right],
\end{equation}

In stage 4, the total loss function is added by $\ell _{1}$ and $\ell _{2}$.

\begin{equation}
\label{eq:Stu_loss}
\ell _{stage4} = \ell _{1} + \ell _{2}.
\end{equation}

\section{Experimental results}

\begin{table*}[t]
	\centering 
	\scriptsize
	\setlength{\tabcolsep}{1.0mm}{
		\begin{tabular}{lccccccccccc}
        \toprule
		\textbf{Model} & \textbf{EN-AR} & \textbf{EN-DE} & \textbf{EN-TR} & \textbf{EN-ES} & \textbf{EN-FR} & \textbf{EN-IT} & \textbf{EN-NL} & \textbf{Avg.} & \textbf{Embedding size} & \textbf{Encoder size} \\ 
		\midrule
		\multicolumn{11}{l}{\emph{Pre-trained Model}}\\
		\verb|mBERT(mean)|  & 16.7 & 33.9 & 16.0 & 21.5 & 33.0 & 34.0 & 35.6 & 27.2 & 92.20M & 85.05M \\ 
		\verb|XLM-R(mean)| & 17.4 & 21.3 & 9.2 & 10.9 & 16.6 & 22.9 & 26.0 & 17.8  & 192.40M & 85.05M  \\ 
		\verb|mBERT-nli-stsb| & 30.9 & 62.2 & 23.9 & 45.4 & 57.8 & 54.3 & 54.1 & 46.9& 92.20M & 85.05M  \\ 
		\verb|XLM-R-nli-stsb| & 44.0 & 59.5 & 42.4 & 54.7 & 63.4 & 59.4 & 66.0 & 55.6 & 192.40M & 85.05M \\
		\verb|LASER| & 66.5 & 64.2 & 72.0 & 57.9 & 69.1 & 70.8 & 68.5 & 67.0 & 23.56M & 17.06M \\
		\verb|LaBSE| & 74.5 & 73.8 & 72.0 & 65.5 & 77.0 &  76.9 & 75.1 & 73.5 & 385.28M & 85.05M \\
		\midrule
		\multicolumn{11}{l}{\emph{Knowledge Distillation}} \\ 
		\verb|mBERT|$\leftarrow$ \verb|SBERT-nli-stsb|  & 77.2 & 78.9 & 73.2 & 79.2 & 78.8 & 78.9 & 77.3 & 77.6 & 92.20M & 85.05M \\ 
		\verb|DistilmBERT|$\leftarrow$ \verb|SBERT-nli-stsb|  & 76.1 & 77.7 & 71.8 & 77.6 & 77.4 & 76.5 & 74.7 & 76.0 & 92.20M & 46.10M \\ 
		\verb|XLM-R|$\leftarrow$ \verb|SBERT-nli-stsb|  & 77.8 & 78.9 & 74.0 & 79.7 & 78.5 & 78.9 & 77.7 & 77.9 & 192.40M & 85.05M\\ 
		
		\verb|mBERT|$\leftarrow$ \verb|SBERT-paraphrases|  & 80.8  & 83.6 & 77.9 & 83.6 & 84.6 & 84.6 & 84.2 & 82.7 & 92.20M & 85.05M\\
		\verb|DistilmBERT|$\leftarrow$ \verb|SBERT-paraphrases|  & 79.7  & 81.7 & 76.4 & 82.3 & 83.2 & 84.3 & 83.0 & 81.5 & 92.20M & 46.10M\\
		
		\verb|XLM-R|$\leftarrow$ \verb|SBERT-paraphrases|  & 82.3  & 84.0 & 80.9 & 83.1 & 84.9 & 86.3 & 84.5 & \textbf{83.7} & 192.40M & 85.05M\\
		\verb|MiniLM|$\leftarrow$ \verb|SBERT-paraphrases|  & 81.3  & 82.7 & 74.8 & 83.2 & 80.3 & 82.4 & 82.2 & \textbf{80.9} & 96.21M & 21.29M\\ 
		\midrule
		\multicolumn{11}{l}{\emph{Ours(Teacher model=SBERT-paraphrases)}} \\ 
		\verb|XLM-R|($b=True$, $bs=128$, $\left | RU \right | = 3$)  & 78.0  & 79.8 & 73.9 & 80.5 & 82.1 & 80.3 & 81.2 & 79.4 & 32.49M & 21.26M\\ 
		\verb|XLM-R|($b=True$, $bs=128$, $\left | RU \right | = 12$)  & 79.4  & 83.6 & 78.7 & 83.3 & 84.2 & 85.6 & 84.8 & \textbf{82.8} & \textbf{32.49M} & 85.05M\\ 
		\verb|XLM-R|($b=False$, $\left | RU \right | = 3$)  & 81.1  & 84.3 & 79.8 & 82.6 & 84.5 & 84.8 & 85.4 & \textbf{83.2} & 192.40M & \textbf{21.26M}\\ 

        \midrule
        \multicolumn{11}{l}{\emph{Ours(Teacher model=SBERT-paraphrases)}} \\ 
		\verb|MiniLM|($b=True$, $bs=128$, $\left | RU \right | = 3$)  & 73.0  & 76.0 & 63.7 & 71.4 & 71.8 & 72.1 & 74.7 & 71.8 & 32.05M & 5.32M\\ 
		\verb|MiniLM|($b=True$, $bs=128$, $\left | RU \right | = 12$)  & 79.7  & 81.0 & 74.1 & 81.9 & 80.1 & 80.8 & 80.7 & \textbf{79.8} & \textbf{32.05M} & 21.29M\\ 
		\verb|MiniLM|($b=False$, $\left | RU \right | = 3$)  & 82.3  & 82.8 & 76.9 & 82.1 & 80.5 & 82.3 & 82.4 & \textbf{81.3} & 96.21M & \textbf{5.32M}\\ 

        \bottomrule

	\end{tabular}
	
	}

	\caption{Spearman rank correlation ($\rho \times 100$) between the cosine similarity of sentence representations and the gold labels for STS 2017-extend \textbf{cross-lingual} dataset. $b$ indicates whether to use Embedding Bottleneck strategy, $bs$ indicates the hidden size of Bottleneck layer. $|RU|$ indicates the first $|RU|$ layers from the basic model are taken as Recurrent Unit, the recurrent times $=$ basic model layers$/|RU|$.}
	\label{table_sts2017_cross}
\end{table*}

\subsection{Evaluation setup}
\label{sec:eval_setup}
\textbf{Dataset}.
The semantic text similarity (STS) task requires models to assign a semantic similarity score between 0 and 5 to a pair of sentences. Following \citet{reimersgurevych2020making}, we evaluate our method on two multilingual STS tasks, i.e., STS2017 \citep{ceretal2017semeval} and STS2017-extend \citep{reimersgurevych2020making}, which contain three monolingual tasks (EN-EN, AR-AR, ES-ES) and six cross-lingual tasks (EN-AR, EN-ES, EN-TR, EN-FR, EN-IT, EN-NL).

\textbf{Parallel corpus}.
In stage 1, stage 2 and stage 3, we use TED2020 \citep{reimersgurevych2020making}, WikiMatrix \citep{schwenketal2021wikimatrix}, Europarl \citep{koehn2005europarl} and NewsCommentary \citep{tiedemann2012parallel} as parallel corpus for training. In stage 4, TED2020 is used for contrastive learning. In this way, the student model first learns generalized multilingual knowledge and then possesses semantic similarity capabilities.

\textbf{Metric}.
Spearman’s rank correlation $\rho$ is reported in our experiments. Specifically, we first compute the cosine similarity score between two sentence embeddings, then calculate the Spearman rank correlation $\rho$ between the cosine score and the golden score.

\subsection{Implementation details}
\label{Implementation}
Mean pooling is applied to obtain sentence embeddings, and the max sequence length is set to 128. We use AdamW \cite{loshchilov2017decoupled} optimizer with a learning rate of 2e-5 and a warm-up of 0.1. In stage1, stage2, and stage3, the models are trained for 20 epochs with a batch size of 64, while in stage 4, the student model is trained for 60 epochs. The mBERT, XLM-R used in this work are base-size model obtained from Huggingface's \textit{transformers} package \cite{wolf2020transformers}, and the MiniLM refers to \textit{MiniLM-L12-H384}\footnote{https://huggingface.co/microsoft/Multilingual-MiniLM-L12-H384}. Our implementation is based on \textit{UER}\cite{zhao2019uer}.

\subsection{Performance comparison}

We compare the model obtained from our multi-stage distillation with the previous state-of-the-art models, and results are shown in Table \ref{table_sts2017_mono} and Table \ref{table_sts2017_cross}. In \textit{Pre-trained Model}, \verb|mBERT(mean)| and \verb|XLM-R(mean)| are mean pooled mBERT and XLM-R models. \verb|mBERT-nli-stsb| and \verb|XLM-R-nli-stsb| are mBERT and XLM-R fine-tuned on the NLI and STS training sets. LASER and LaBSE are obtained from \citet{artetxeschwenk2019massively} and \citet{feng2020language}. In \textit{Knowledge Distillation}, we use the paradigm of \verb|Student|$\leftarrow$\verb|Teacher| to represent the \verb|Student| model distilled from the \verb|Teacher| model. There are two teacher models, i.e., \verb|SBERT-nli-stsb| and \verb|SBERT-paraphrases|, which are released by UKPLab\footnote{https://github.com/UKPLab/sentence-transformers}. The former is fine-tuned on the English NLI and STS training sets, and the latter is trained on more than 50 million English paraphrase pairs. The student models include mBERT, XLM-R, DistilmBERT \citep{sanh2019distilbert} and MiniLM \citep{wangetal2021minilmv2}.

Table \ref{table_sts2017_mono} and Table \ref{table_sts2017_cross} show the evaluation results on monolingual and multilingual STS task, respectively. For the XLM-R, our method compresses the embedding size by 83.2\% with 0.3\% worse monolingual performance and 0.9\% worse cross-lingual performance, compresses the encoder size by 75\% with slightly higher (0.4\%) monolingual performance and 0.5\% worse cross-lingual performance. When compressing the embedding layer and the encoder simultaneously, the model size is reduced by 80.6\%, its monolingual performance drop by 2\% and cross-lingual performance drop by 4\%, but it still outperforms the pre-trained models.

For comparison with other distillation methods, MiniLM$\leftarrow$ SBERT-paraphrases is taken as a strong baseline. Our framework can further compress its embedding size by 66.7\% with 0.6\% worse in monolingual performance and 1.1\% worse in cross-lingual performance. Its encoder size is further compressed by 75\% with slightly higher monolingual (0.1\%) and cross-lingual (0.4\%) performance. In addition, our compressed XLM-R($b=True$, $bs=128$, $|RU| = 12$) achieves higher monolingual(0.8\%) and cross-lingual(1.9\%) performance with the same model size.

\subsection{Ablation study}

\begin{table}[tb]
	\centering
	\scriptsize
	\setlength{\tabcolsep}{3.5mm}{

	\begin{tabular}{lcccc}
		\toprule
		\textbf{Model} &  \textbf{AR-AR} & \textbf{ES-ES} & \textbf{EN-EN}   & \textbf{Avg.} \\ 
		\midrule
		ours & 76.7 & 84.5 & 86.6 & 82.6 \\ 
		w/o MCL & 76.4 & 83.9 & 86.8 & 82.3 \\ 
		w/o Rec. & 67.4  & 80.1 & 86.6 & 78.0 \\ 
		w/o MCL+Rec. & 67.9 & 79.3 & 86.6 & 77.9 \\ 
		\bottomrule
	\end{tabular}
	}
	\caption{Results of ablation studies on STS-2017 monolingual task}
	\label{table_ablation_sts2017}
\end{table}

\begin{table}[tb]
	\centering 
	\scriptsize
	\setlength{\tabcolsep}{0.3mm}{
	\begin{tabular}{lccccccccc}
		\toprule
		\textbf{Model} & \textbf{EN-AR} & \textbf{EN-DE} & \textbf{EN-TR} & \textbf{EN-ES} & \textbf{EN-FR} & \textbf{EN-IT} & \textbf{EN-NL} & \textbf{Avg.} \\ 
		\midrule
		\verb|ours|  & 78.0  & 79.8 & 73.9 & 80.5 & 82.1 & 80.3 & 81.2 & 79.4 \\ 
		w/o MCL & 75.9 & 79.7 & 73.2 & 79.9 & 80.4 & 80.4 & 80.5 & 78.5  \\ 
		w/o Rec. & 69.1 & 73.4 & 66.5 & 70.2 & 73.7 & 73.0 & 75.9 & 71.7 \\ 
		w/o MCL+Rec. & 67.8 & 73.6 & 66.4 & 68.5 & 72.8 & 71.8 & 75.2 & 70.9 \\
		\bottomrule
	\end{tabular}
	}
	\caption{Results of ablation studies on STS2017-extend cross-lingual task}
	\label{table_ablation_cross}
\end{table}

Among the three key strategies, multilingual contrastive learning (\textbf{MCL}) and parameter recurrent (\textbf{Rec.}) are two crucial mechanisms to improve model performance. The bottleneck is used to compress the model. In this section, ablation studies is performed to investigate the effects of \textbf{MCL} and \textbf{Rec.}. The effects of the bottleneck will be discussed in section \ref{sec:deepvswidth}.

 XLM-R($b$=True, $bs$=128, $|RU|$ = 3) is selected as the basic model. We consider three different settings: 1) training without MCL task. 2) training without parameter recurrent. 3) training without both. The monolingual results and multilingual results are presented in Table \ref{table_ablation_sts2017} and Table \ref{table_ablation_cross}.

It can be observed that: 1) without MCL task, the model performs poorer on the cross-lingual tasks. 2) without parameter sharing, the model performs poorer on all datasets. 3) MCL task can significantly improve the cross-lingual performance on EN-AR, EN-ES, EN-FR, EN-NL. It can be concluded that both MCL task and parameter recurrent play a key role in our method.

\begin{figure}[t]
	\centering
	\includegraphics[width=0.9\linewidth]{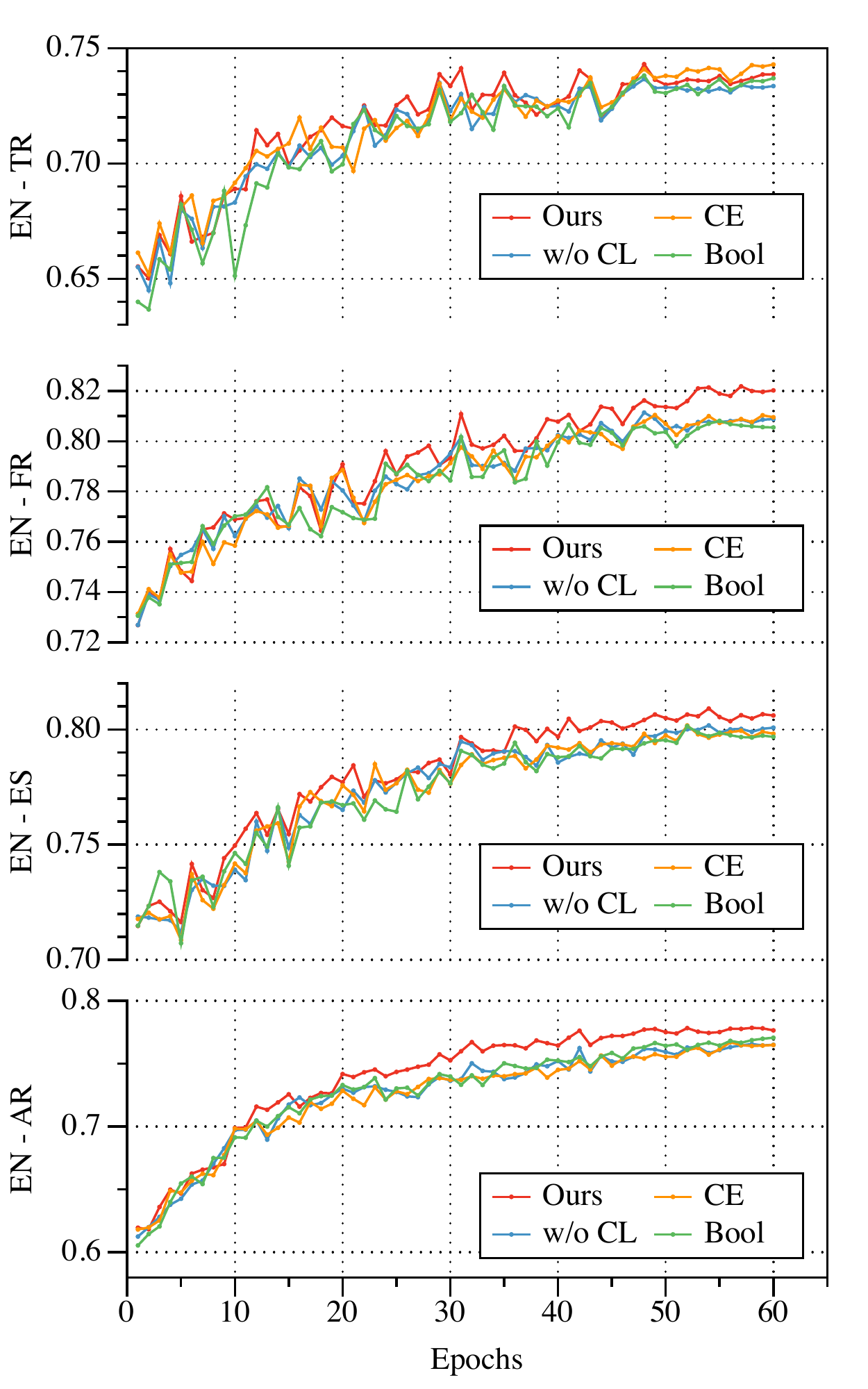}
	\caption{Performance of XLM-R ($b$=True, $bs$=128, $|RU|$ = 3) after each training epoch on EN-AR, EN-ES, EN-FR, EN-TR tasks with different contrastive learning settings. }
	\label{fig:epoch}
\end{figure}

\begin{table}[t]
	\centering 
	\scriptsize
	\setlength{\tabcolsep}{0.5mm}{
	\begin{tabular}{lccccccccc}
		\toprule
		\textbf{Settings} & \textbf{EN-AR} & \textbf{EN-DE} & \textbf{EN-TR} & \textbf{EN-ES} & \textbf{EN-FR} & \textbf{EN-IT} & \textbf{EN-NL} & \textbf{Avg.} \\ 
		\midrule
		\verb|Ours|  & 78.0  & 79.8 & 73.9 & 80.5 & 82.1 & 80.3 & 81.2 & 79.4 \\ 
		\textit{Bool}  & 77.0$\downarrow$  & 80.5$\uparrow$ & 73.5$\downarrow$ & 79.8$\downarrow$ & 80.3$\downarrow$ & 80.7$\uparrow$ & 81.2 & 79.0$\downarrow$ \\ 
		\textit{CE}  & 76.6$\downarrow$ & 79.9$\uparrow$ & 74.3$\uparrow$ & 80.0$\downarrow$ & 80.8$\downarrow$ & 80.6$\uparrow$ & 80.7$\downarrow$ & 78.9$\downarrow$ \\
		\textit{w/o CL} & 75.9$\downarrow$ & 79.7$\downarrow$ & 73.2$\downarrow$ & 79.9$\downarrow$ & 80.4$\downarrow$ & 80.4$\uparrow$ & 80.5$\downarrow$ & 78.5$\downarrow$ \\ 
		\bottomrule
	\end{tabular}
	}
	\caption{Evaluation results of XLM-R ($b=True$, $bs=128$, $\left | RU \right | = 3$) on the STS2017-extend cross-lingual task with different contrastive learning settings.}
		\label{table_mcl_across}
\end{table}

\subsection{Effect of contrastive learning}

\begin{table}[tb]
	\centering 
	\scriptsize
	\setlength{\tabcolsep}{2mm}{
	\begin{tabular}{lcc}
		\toprule
		\textbf{Settings} & \textbf{Avg. (Monolingual)} & \textbf{Avg. (Cross-lingual)}\\ 
		\midrule
          \multicolumn{3}{l}{\emph{Single-stage}} \\ 
  		Random Initialize & 78.1 & 71.1 \\ 
		\quad + Pre-Distillation & 79.0 & 73.8 \\ 
		\midrule
        \multicolumn{3}{l}{\emph{Multi-stage}} \\
		stage 1 + 2  & 48.4  & 20.8   \\ 
		stage 1 + 2 + 3 & 75.2 & 70.6 \\
		stage 1 + 2 + 3 + 4 & \textbf{82.6} & \textbf{79.4} \\
		\bottomrule
	\end{tabular}
	}
	\caption{Comparison of using different stage settings on monolingual and multilingual STS task. XLM-R is the basic model. The first three layers from XLM-R are taken as a Recurrent Unit, bottleneck hidden size is 128.}
	\label{table_multistage}
\end{table}

\begin{table*}[t]
	\centering 
	\scriptsize
	\setlength{\tabcolsep}{4mm}{
		\begin{tabular}{lcccccc}
        \toprule
		\textbf{Model} & \textbf{AR-AR} & \textbf{ES-ES} & \textbf{EN-EN}  & \textbf{Avg.} & \textbf{Embedding size} & \textbf{Encoder size} \\ 
		\midrule
		\multicolumn{6}{l}{\emph{Teacher model=SBERT-paraphrases, Student model=XLM-R, $\left | RU \right | = 3$}}\\
		$b=True$, $bs=128$  & 76.7  & 84.5 & 86.6  & \textbf{82.6} & 32.49M & 21.26M\\ 
		$b=True$, $bs=256$  & 76.2  & 84.9 & 87.4 & \textbf{82.8} & 64.59M & 21.26M\\ 
		$b=False$ & 79.9  & 86.8 & 88.4 & \textbf{85.0} & 192.40M & 21.26M\\ 

		\midrule
		\multicolumn{6}{l}{\emph{Teacher model=SBERT-paraphrases, Student model=XLM-R, $b=True$, $bs=128$ }}\\
	 	$\left | RU \right | = 3$  & 76.7  & 84.5 & 86.6  & \textbf{82.6} & 32.49M & 21.26M\\
 		$\left | RU \right | = 6$ & 78.1  & 84.8 & 87.4 & \textbf{83.4} & 32.49M & 42.52M\\
		$\left | RU \right | = 12$  & 79.0  & 85.5 & 88.4 & \textbf{84.3} & 32.49M & 85.05M\\

        \midrule
        \multicolumn{6}{l}{\emph{Teacher model=SBERT-paraphrases, Student model=MiniLM, $\left | RU \right | = 3$}} \\ 
		$b=True$, $bs=128$ & 72.8  & 79.3 & 84.4 & \textbf{78.8} & 32.05M & 5.32M\\ 
		$b=True$, $bs=256$  & 72.2  & 81.2 & 85.2 & \textbf{79.5} & 64.10M & 5.32M\\ 
		$b=False$ & 79.9  & 85.3 & 85.6 & \textbf{83.6} & 96.21M & 5.32M\\ 
		\midrule
		\multicolumn{6}{l}{\emph{Teacher model=SBERT-paraphrases, Student model=MiniLM, $b=True$, $bs=128$ }}\\
		$\left | RU \right | = 3$ & 72.8  & 79.3 & 84.4 & \textbf{78.8} & 32.05M & 5.32M\\ 
		$\left | RU \right | = 6$  & 75.6  & 83.8 & 85.1 & \textbf{81.5} & 32.05M & 10.64M\\ 
	    $\left | RU \right | = 12$  & 79.0  & 84.4 & 85.2 & \textbf{82.9} & 32.05M & 21.29M\\ 

        \bottomrule

	\end{tabular}
	
	}

	\caption{The performance of STS2017 \textbf{monolingual} task based on XLM-R($b$=True, $bs$=128, $|RU|$ = 3) and MiniLM($b$=True, $bs$=128, $|RU|$ = 3), We evaluated the effect of increasing $bs$ or $|RU|$.}
	\label{table_deepvswidth_mono}
\end{table*}

\begin{table*}[t]
	\centering 
	\scriptsize
	\setlength{\tabcolsep}{1.0mm}{
		\begin{tabular}{lccccccccccc}
        \toprule
		\textbf{Model} & \textbf{EN-AR} & \textbf{EN-DE} & \textbf{EN-TR} & \textbf{EN-ES} & \textbf{EN-FR} & \textbf{EN-IT} & \textbf{EN-NL} & \textbf{Avg.} & \textbf{Embedding size} & \textbf{Encoder size} \\ 
		
        \midrule
        \multicolumn{11}{l}{\emph{Teacher model=SBERT-paraphrases, Student model=XLM-R, $\left | RU \right | = 3$ }} \\ 
    	$b=True$, $bs=128$  & 78.0  & 79.8 & 73.9 & 80.5 & 82.1 & 80.3 & 81.2 & \textbf{79.4} & 32.49M & 21.26M\\ 
		$b=True$, $bs=256$  & 79.2  & 81.8 & 73.8 & 82.3 & 82.7 & 81.6 & 82.6 & \textbf{80.6} & 64.59M & 21.26M\\ 
		$b=False$  & 81.1  & 84.3 & 79.8 & 82.6 & 84.5 & 84.8 & 85.4 & \textbf{83.2} & 192.40M & 21.26M\\
		  
        \midrule
        \multicolumn{11}{l}{\emph{Teacher model=SBERT-paraphrases, Student model=XLM-R, $b=True$, $bs=128$ }} \\ 
		
	    $\left | RU \right | = 3$   & 78.0  & 79.8 & 73.9 & 80.5 & 82.1 & 80.3 & 81.2 & \textbf{79.4} & 32.49M & 21.26M\\ 
	    $\left | RU \right | = 6$   & 78.8  & 80.0 & 74.7 & 82.9 & 83.5 & 83.4 & 84.6 & \textbf{81.1} & 32.49M & 42.52M\\ 
	   	$\left | RU \right | = 12$  & 79.4  & 83.6 & 78.7 & 83.3 & 84.2 & 85.6 & 84.8 & \textbf{82.8} & 32.49M & 85.05M\\ 
	
		\midrule
        \multicolumn{11}{l}{\emph{Teacher model=SBERT-paraphrases, Student model=MiniLM, $\left | RU \right | = 3$ }} \\ 
		   
		$b=True$, $bs=128$ & 73.0  & 76.0 & 63.7 & 71.4 & 71.8 & 72.1 & 74.7 & \textbf{71.8} & 32.05M & 5.32M\\ 
		$b=True$, $bs=256$ & 69.7  & 77.1 & 66.2 & 73.5 & 73.5 & 74.3 & 75.6 & \textbf{72.8} & 64.10M & 5.32M\\ 
		$b=False$  & 82.3  & 82.8 & 76.9 & 82.1 & 80.5 & 82.3 & 82.4 & \textbf{81.3} & 96.21M & 5.32M\\ 

		\midrule
		\multicolumn{11}{l}{\emph{Teacher model=SBERT-paraphrases, Student model=MiniLM, $b=True$, $bs=128$}} \\ 
		$\left | RU \right | = 3$ & 73.0  & 76.0 & 63.7 & 71.4 & 71.8 & 72.1 & 74.7 & \textbf{71.8} & 32.05M & 5.32M\\ 
		$\left | RU \right | = 6$ & 77.1  & 78.7 & 68.2 & 78.1 & 75.9 & 77.0 & 77.6 & \textbf{76.1} & 32.05M & 10.64M\\ 
        $\left | RU \right | = 12$  & 79.7  & 81.0 & 74.1 & 81.9 & 80.1 & 80.8 & 80.7 & \textbf{79.8} & 32.05M & 21.29M\\ 
        
        \bottomrule

	\end{tabular}
	
	}

	\caption{The performance of STS2017-extend \textbf{cross-lingual} task based on XLM-R($b$=True, $bs$=128, $|RU|$ = 3) and MiniLM($b$=True, $bs$=128, $|RU|$ = 3), We evaluated the effect of increasing $bs$ or $|RU|$.}
	\label{table_deepvswidth_cross}
\end{table*}

To investigate the effects of contrastive learning in stage 4, we select XLM-R($b$=True, $bs$=128, $|RU|$ = 3), modify the original objective in (\ref{eq:MCL}) into three different settings, namely, \textit{Bool}, \textit{CE} and \textit{w/o CL}.  

In the \textit{Bool} setting, the soft label in (\ref{eq:MCL}) is replaced with hard label (0 or 1), as (\ref{eq:01mse}), 
\begin{equation}
\label{eq:01mse}
\ell _{Bool}  = \frac{1}{N^{2} } \sum_{i}^{N}\sum_{j}^{N}  \left (\delta  (h_{T}^{si},h_{T}^{sj}) -\phi (h_{S}^{si},h_{S}^{tj})  \right ) ^{2},
\end{equation}
where $\delta(x,y) = 1$, if $x=y$, otherwise $0$.

In the \textit{CE} setting, the objective in (\ref{eq:MCL}) is replaced with temperature-scaled cross-entropy, as (\ref{eq:kld}),
\begin{align}
\label{eq:kld}
    \ell _{CE} = - \sum_{i}^{N}\sum_{j}^{N}\phi_{T}\log \frac{e^{\phi_{S}/ \tau }}{\sum_{k=1}^{N}{e^{\phi_{S}/\tau}}},
\end{align}
where $\phi_T=cos(h_{T}^{si}, h_{T}^{sj})$, $ \phi_S=cos(h_{S}^{si} , h_{S}^{tj})$, $\tau=0.05$ is a hyperparameter called temperature.

In the \textit{w/o CL} setting, the contrastive learning is removed in Stage 4.

Table \ref{table_mcl_across} presents the model performance of cross-lingual semantic similarity task with different settings. It can be observed that all the above training objectives can improve the model performance on the cross-lingual task, compared with the \textit{w/o CL} settings. Model trained with (\ref{eq:01mse}) and (\ref{eq:kld}) underperform that trained with (\ref{eq:MCL}), especially on EN-AR, EN-ES, EN-FR, EN-NL task.

We plot the convergence process of different settings in Figure \ref{fig:epoch}. On EN-AR, EN-ES, EN-FR tasks, our setting outperform other settings. It is worth mentioning that on the EN-TR task, our setting underperform the \textit{CE} setting according to Table \ref{table_mcl_across}. However, our setting reaches the same level as \textit{CE} setting during the 30 to 40 epoch.

\subsection{Effect of multi-stages}
To verify the effectiveness of multi-stages, we shows the performance comparison of using different stage settings in Table \ref{table_multistage}. In the \textit{Single-stage} setting, we first initialize the shrunk student model in two ways: (1) Random Initialize: Adding the untrained embedding bottleneck layers to the student model. (2) Pre-Distillation: The student model with bottleneck layer is initialized by distillation using XLM-R and the same corpus as section \ref{sec:eval_setup}. Then we follow  \citet{reimersgurevych2020making} to align vector space between different languages. In the \textit{Multi-stage} setting, the performance of the student model is reported after each stage.

As shown in Table \ref{table_multistage}, the \textit{Multi-stage} setting outperforms the single-stage one, indicating that our multi-stage framework with an assistant model is effective. Adding stage3 and stage4 further improves the student model performance, suggesting that multi-stage training are necessary.

\subsection{Effect of bottleneck and recurrent unit}
\label{sec:deepvswidth}

In this section, we study the impact of embedding bottleneck and recurrent unit strategies on multilingual semantic learning. We consider three settings for each strategy, as shown in Table \ref{table_deepvswidth_mono} and Table \ref{table_deepvswidth_cross}.

First, we found that both XLM-R and MiniLM perform better as the bottleneck hidden size $bs$ increases. The performance is best when the entire embedding layer is retained, The MiniLM($b$=False) can outperform its original model in Table \ref{table_sts2017_mono} and Table \ref{table_sts2017_cross}. But the benefit of increasing $bs$ is not obvious unless the entire embedding layer is retained.

Second, by increasing the number of recurrent unit layers $|RU|$, XLM-R and MiniLM have been steadily improved on these two tasks. The increase in model size caused by the $|RU|$ is less than the $bs$. For example, the performance of MiniLM on cross-lingual tasks increased by 8\%, while its size only increased by 15.9M.

Finally, it can be observed that when using the bottleneck layer ($b$=True), the model performance will increase steadily as $|RU|$ increases. The smaller the encoder hidden size, the more significant effect caused by $|RU|$ increasing ($\Delta$MiniLM>$\Delta$XLM-R). However, the increase of $bs$ can not improve performance significantly but make the embedding size larger. Therefore, an effective way to compress the multilingual model is reducing $bs$ while increasing $|RU|$. In this way, we shrink XLM-R by 58\%, MiniLM by 55\%, with less than 1.1\% performance degradation.

\section{Conclusion}

In this work, we realize that the cross-lingual similarity matching task requires a large model size. To obtain a small-size model with cross-lingual matching ability, we propose a multi-stage distillation framework. Knowledge distillation and contrastive learning are combined in order to compress model with less semantic performance loss.

Our experiments demonstrate promising STS results with three monolingual and six cross-lingual pairs, covering eight languages. The empirical results show that our framework can shrink XLM-R or MiniLM by more than 50\%. In contrast, the performance is only reduced by less than 0.6\% on monolingual and 1.1\% on cross-lingual tasks. If we slack the tolerated loss performance in 4\%, the size of XLM-R can be reduced by 80\%.

\bibliography{anthology,custom}
\bibliographystyle{acl_natbib}

\end{document}